\documentclass[oneside,english]{article}
\usepackage[margin=1in]{geometry}

\usepackage{xspace}
\usepackage{booktabs}
\usepackage{authblk}
\usepackage{amsmath}
\usepackage{amsfonts}
\usepackage{graphicx}
\usepackage{cleveref}
\usepackage{natbib}

\newcommand*{\Reals}{\ensuremath{\mathbb{R}}}
\newcommand*{\PReals}{\ensuremath{\mathbb{R}_{>0}}}
\newcommand{\Distr}{\operatorname{Distr}}
\newcommand{\hatS}{\hat{S}}
\newcommand{\hatH}{\hat{H}}
\newcommand{\hath}{\hat{h}}

\newcommand{\calS}{\mathcal{S}}
\newcommand{\calT}{\mathcal{T}}

\newcommand{\EE}{\ensuremath{\mathbb{E}}}
\newcommand{\II}{\ensuremath{\mathbb{I}}}

\newcommand{\See}[2][]{(\Cref{#2}#1)\xspace}
\newcommand{\pkg}[1]{\textbf{#1}\xspace}
\newcommand{\code}[1]{\texttt{#1}\xspace}
\newcommand{\Rstats}{\textsf{R}\xspace}

\begin{document}

\title{Avoiding C-hacking when evaluating survival distribution predictions with discrimination measures}

\author[1,2]{Raphael Sonabend\thanks{raphaelsonabend@gmail.com}}
\author[3]{Andreas Bender}
\author[2,4,5]{Sebastian Vollmer}

\affil{MRC Centre for Global Infectious Disease Analysis, Jameel Institute, Imperial College London, School of Public Health, W2 1PG, London, UK}
\affil[2]{Department of Computer Science, Technische Universität Kaiserslautern, Gottlieb-Daimler-Straße 47, 67663, Kaiserslautern, Germany}
\affil[3]{Department of Statistics, LMU Munich, Ludwigstr. 33, 80539, Bavaria, Germany}
\affil[4]{Data Science and its Application, Deutsches Forschungszentrum für Künstliche Intelligenz (DFKI), Trippstadter Str. 122, 67663, Kaiserslautern, Germany}
\affil[5]{
Mathematics Institute, University of Warwick, Zeeman Building, CV4 7AL, Coventry, UK}

\thispagestyle{empty}
\maketitle

\abstract{In this paper we consider how to evaluate survival distribution predictions with measures of discrimination. This is a non-trivial problem as discrimination measures are the most commonly used in survival analysis and yet there is no clear method to derive a risk prediction from a distribution prediction. We survey methods proposed in literature and software and consider their respective advantages and disadvantages. Whilst distributions are frequently evaluated by discrimination measures, we find that the method for doing so is rarely described in the literature and often leads to unfair comparisons. We find that the most robust method of reducing a distribution to a risk is to sum over the predicted cumulative hazard. We recommend that machine learning survival analysis software implements clear transformations between distribution and risk predictions in order to allow more transparent and accessible model evaluation. The code used in the final experiment is available at https://github.com/RaphaelS1/distribution\_discrimination.}

\newpage

\section{Introduction}
Predictive survival models estimate the distribution of the time until an event of interest takes place. This prediction may be presented in one of three ways, as a: i) time-to-event, $Y \in \PReals$, which represents the time until the event takes place; ii) a relative risk, $\phi \in \Reals$, which represents the risk of the event taking place compared to other subjects in the same sample; or iii) the probability distribution for the time to the event, $\calS \in \Distr(\PReals)$, where $\Distr(\PReals)$ is the set of distributions over $\PReals$. Less abstractly, consider the Cox PH \citep{Cox1972}: $h(t) = h_0(t)\exp(X\beta)$ where $h_0$ is the `baseline' hazard function, $X$ are covariates, and $\beta$ are coefficients to be estimated. In practice, software fits the model by estimating the coefficients, $\hat{\beta}$. Predictions from the fitted model may then either be returned as a relative risk prediction, $X\hat{\beta}$ or $\exp(X\hat{\beta})$, or $h_0$ is also estimated and a survival distribution is predicted as $\hath(t) = \hath_0(t)\exp(X\hat{\beta})$.

The Cox PH is a special type of survival model that can naturally return both a survival distribution and a relative risk prediction, however this is not the case for all models. For example, random survival forests \citep{Ishwaran2008} only return distribution predictions by recursively splitting observations into increasingly homogeneous groups and then fitting the Nelson-Aalen estimator in the terminal node.
\\\\
The most common method of evaluating survival models is with discrimination measures \citep{Collins2014, Gonen2005, Rahman2017}, in particular Harrell's \citep{Harrell1982} and Uno's C \citep{Uno2011}. These measures determine if relative risk predictions are concordant with the true event time. To give a real-world example, a physician may predict that a 70-year old patient with cancer is at higher risk of death than a 12-year old patient with a broken arm. If the 70-year old dies before the 12-year old then the risk prediction is said to be concordant with the observed event times as the patient with the predicted higher risk died first.

Despite there being no `obvious' method of evaluating discrimination from a distribution prediction, papers that compare (or benchmark) model discrimination, frequently omit stating the software and/or method used for evaluating distribution predictions (e.g. \cite{Fernandez2016, Herrmann2020, Spooner2020, Zhang2021}).

In this paper, we will consider how to evaluate methods, which only predict survival distributions, with measures of discrimination. We will consider these methods in the context of model comparison, i.e. by establishing if the measure accurately evaluates if one model can be said to be superior to another. For example, if a given random survival forest (distribution prediction only) has better discrimination than a support vector machine (risk prediction only) \cite{VanBelle2007}.
\\\\
We will review methods of discrimination evaluation that are discussed in the literature and discuss their advantages and disadvantages. First some notation to be used throughout the paper: let $X_i \in \Reals^p$ be $p$ covariates for subject $i$, let $Y_i$ be the true (but unobserved) survival time; $C_i$ be the true (but unobserved) censoring time, and $T_i = \min(Y_i, C_i)$ be the observed outcome time; finally let $\Delta_i = \II(T_i = Y_i)$ be the survival indicator. In this paper we do not consider competing risks settings, which require specialised measures.

\section{Methods}

We consider how discrimination measures are utilised in the literature to evaluate the predictive performance of models that predict survival distributions \See{sec:lit}, we then review the identified methods (Sections \ref{sec:timedep}-\ref{sec:timeind}). To illustrate our findings we provide a worked example in \Cref{sec:ex}. The focus in our review is not to compare the (dis)advantages of measures but instead their compatibility. For example, we do not compare if Antolini's C is `better' than Harrell's C but instead note that the former requires a distribution prediction and the latter a risk prediction, hence estimated C-statistics cannot be compared between measures as they evaluate separate predictions types (we call such a comparison `C-hacking').
 
\subsection{Literature Review}
\label{sec:lit}

We first performed a formal literature review using PubMed and then a less formal review from articles and software packages that had been drawn to our attention. The purpose of the review was to determine how model discrimination predictions have historically been evaluated for machine learning models that make distributional predictions.

\begin{figure}
    \centering
    \includegraphics[width=8cm]{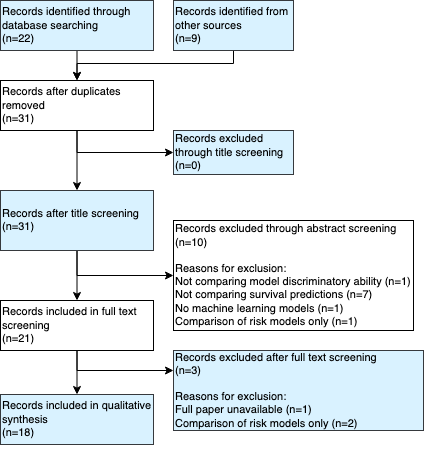}
    \caption{PRISMA diagram for literature review. Inclusion criteria: Articles that compare machine learning survival predictions with measures of discrimination.}
    \label{fig:prisma}
\end{figure}

We searched PubMed for `(comparison OR benchmark) AND ("survival analysis" OR "time-to-event analysis") AND "machine learning" AND (discrimination OR concordance OR "C statistic" OR "c index")'. We excluded articles if: i) they did not use measures of discrimination; ii) no machine learning models were included; iii) only risk-prediction models were included; iv) the models did not make survival predictions (e.g. classifiers). We found 22 articles in our initial search, which were reduced to nine after screening, a full PRISMA diagram is provided in \Cref{fig:prisma}, the diagram includes nine other records which were identified outside of the search and which are discussed further below.
\\\\

A total of nine articles were retained for qualitative assessment \citep{Hadanny2022,Johri2021,Loureiro2021,MosqueraOrgueira2020,Aivaliotis2021,Kantidakis2020,Spooner2020,Crombe2021,Herrmann2020}. All of these, without exception, compared risk-predicting Cox-based models (e.g. regularised, boosted, neural adaptations) to random survival forests \cite{Ishwaran2008}. \pkg{scikit-survival}\citep{pkgsksurvival}, \pkg{randomForestSRC}\citep{pkgrfsrc}, \pkg{ranger}\citep{pkgranger} and \pkg{mlr}\citep{pkgmlr} were utilised to implement and evaluate the RSFs. RSFs make distributional predictions by ensembling a Nelson-Aalen estimator across bootstrapped models \cite{Ishwaran2008}. Transformation from distribution to risk is handled in \pkg{randomForestSRC} and \pkg{scikit-survival} by taking the sum over the predicted cumulative hazard function for each observation, which is recommended by \cite{Ishwaran2008}, we refer to this transformation as `expected mortality' \See{sec:mortality}. In contrast, no transformation is provided in \pkg{ranger}, which only returns a distribution prediction, however this is handled in \cite{Spooner2020} by utilising \pkg{mlr}, which provides the same expected mortality transformation.
\\\\
As well as these articles, we were also aware of the following articles and software that discuss the discrimination of models that make distributional predictions: \cite{Kvamme2019a, Lee2018, Gensheimer2019, Kvamme2019, mlr3proba, Zhao2020, Haider2020, pkgpec, Schwarzer2000}. Of these articles the methods of comparing predicted distribution discriminatory ability are: 1) utilising time-dependent concordance indices \citep{Kvamme2019, Kvamme2019a, Lee2018} \See{sec:timedep}; 2) comparing predicted probabilities at a given time-point \citep{Gensheimer2019, pkgpec, Schwarzer2000, Zhao2020, Zhong2019} \See{sec:improper}; 3) calculating and comparing a summary statistic (e.g. expected survival time) from the predicted distributions \citep{mlr3proba, Haider2020} \See{sec:summary}.
\\\\
These four methods are grouped according to the required measure, i,e.: A) time-dependent discrimination; and B) time-independent discrimination. Discussion follows after defining some useful notation.

In practice, software for time-to-event predictions will usually return a matrix of survival probabilities. Let $[T_0, T_N]$ be the range of observed survival times in a training dataset, let $M$ be the number of observations in the test dataset and let $M^* \leq M$ be the number of time-points for which predictions are made, then we predict $\mathbf{S} \in [0, 1]^{M \times M^*}$, which correspond to predictions of individual survival functions, $S_i(T), T \in \calT \subseteq [T_0, T_N]$.

\subsection{Time-dependent discrimination}
\label{sec:timedep}

Discrimination measures can be computed as the proportion of concordant pairs over comparable pairs. Let $i \neq j$ be a pair of observations with observed outcomes and predicted risks of $\{(T_i, \Delta_i, \phi_i), (T_j, \Delta_j, \phi_j)\}$ respectively. Then $(i,j)$ are comparable if $T_i < T_j \cap \Delta_i$ and the predicted risks are concordant with the outcome times if $\phi_i > \phi_j$. In this paper we are concerned with how the values of $(\phi_i,\phi_j)$ are calculated.
\\\\
Time-dependent discrimination measures define concordance over time either by taking $\phi_i$ to be predicted survival probabilities such as \cite{Antolini2005}, or as predicted linear predictors, such as \cite{Heagerty2000}.

\cite{Antolini2005} define a pair of observations as concordant if the predicted survival probabilities are concordant at the shorter outcome time,

\begin{equation}
    P(\hatS_i(T_i) < \hatS_j(T_i) | T_i < T_j \cap \Delta_i = 1)
\end{equation}

In contrast, \cite{Heagerty2000} calculate the AUC by integrating over specificity and sensitivity measures given by

\begin{align}
    \text{sensitivity}(c, t) = P(\phi_i > c | T_i \leq t) \\
    \text{specificty}(c, t) = P(\phi_i \leq c | T_i > t)
\end{align}

Various metrics have been based on these ideas and several are implemented in the \Rstats package \pkg{survAUC}\citep{pkgsurvauc}. However, all require a single relative risk predictor, and therefore require some transformation from a survival distribution prediction, and secondly all assume a one-to-one relationship between the predicted value and expected survival times (which is unlikely in complex machine learning models), for example a proportional hazards assumption where the predicted risk is related to the predicted survival distribution by multiplication of a constant \citep{pkgsurvauc}.

We are unaware of any time-dependent AUC metrics, except for Antolini's, that evaluates survival time predictions without a further transformation being required. This may explain why Antolini's C-index is seemingly more popular in the artificial survival network literature \citep{Kvamme2019, Kvamme2019a, Lee2018}.

On the surface, time-dependent discrimination measures are optimal for evaluating distributions by discrimination. However, they are a poor choice for model comparison. Time-dependent measures that evaluate risk predictions (such as Heagerty's) require a transformation from survival distribution predictions and any such transformation is unlikely to result in the one-to-one mapping required by the measures. In contrast, Antolini's C evaluates the concordance of a distribution, which means that it can only be used to compare the concordance of two models that make distribution predictions, as opposed to, say, one model that predicts distributions (e.g. RSFs) and one that predicts relative risks (e.g. SVMs). The experiment in \Cref{sec:ex} demonstrates why results from Antolini's C cannot be simply compared to results from other concordance indices.

\subsection{Time-independent discrimination}
\label{sec:timeind}

Time-independent discrimination measures for survival analysis evaluate relative risk predictions by estimating concordance. 

Let $\calS \subseteq \Distr(\PReals)$ be a convex set of distributions over the positive Reals; then we define a \textit{distribution reduction method} as any function of the form: $f: \calS \rightarrow \Reals$, which map a survival distribution prediction, $\zeta \in \calS$, to a single relative risk, $\phi \in \Reals$.  In the discrete software analogue, we consider functions $f': [0,1]^{M^*} \rightarrow \Reals$.

Distribution reduction methods are required to utilise time-independent discrimination measures for models that make distribution predictions. We consider the three from the literature review in turn.

\subsubsection{Comparing probabilities}
\label{sec:improper}
Evaluating discrimination at a given survival time is formally defined by estimating

\begin{equation}
    P(\hatS_i(t) < \hatS_j(t) | T_i < T_j \cap \Delta_i = 1)
\end{equation}

for some chosen $t \in \PReals$. The distribution is reduced to a relative risk by evaluating the survival probabilities at a given time-point, $\phi := \hatS(t')$ where $\hatS$ is the predicted survival function and $t' \in \PReals$. Note the key difference between this method and Antolini's C is that $t$ can be arbitrarily chosen here, whereas Antolini's C estimates the concordance at the observed outcome time.

This method assesses how well a model separates patients at a single time-point; it has several problems: 1) it is not `proper' in the sense that the optimal model may not maximise the concordance at $t'$ \citep{Blanche2019}; 2) it is prone to manipulation as one could select the $t'$ that maximises the C-index for their chosen model (see \Cref{sec:ex}); and 3) if predicted survival curves overlap then evaluation at different time-points will lead to contradictory results (as the observed event time will always stay the same). The above issues apply even if evaluated at several time-points.

\subsubsection{Distribution summary}
\label{sec:summary}
The distribution summary statistic method reduces a probability distribution prediction to a summary statistic, most commonly, the mean or median of the distribution, i.e.,

\begin{align}
    P(\EE[\zeta_i] < \EE[\zeta_j] | T_i < T_j \cap \Delta_i = 1) \\
    P(m(\zeta_i) < m(\zeta_j) | T_i < T_j \cap \Delta_i = 1)
\end{align}

where $m(\zeta_i)$ is the median of distribution $\zeta_i$. In theory, this should provide the most meaningful reduction with a natural interpretation (mean or median survival time), however this is not the case as the presence of censoring means that the predicted survival predictions will usually result in `improper predictions', i.e. the basic properties of the survival function are not satisfied: $\lim_{t \rightarrow +\infty} S_T(t) \neq 0$. To see why this is the case, note that the majority of survival distribution predictions make use of a discrete estimator such as the Kaplan-Meier estimator, which is defined as follows:

\begin{equation}
\hat{S}(t)=
    \begin{cases} 
        1 & t<t_{(1)}\\
        \prod_{i:t_{(i)}\leq t} \left(1-d_i/n_i\right) & t\geq t_{(1)}
    \end{cases}
\end{equation}
where $d_i, n_i$ are the number of deaths and events (death or censoring) at ordered events times time $t_{(i)}, i=1,\ldots,n$. 
By definition of this estimator, unless all observations at risk in the final time-point experience the event ($d_i = n_i$), the predicted survival probability in this last point will be non-zero.

\begin{figure}
    \centering
    \includegraphics[width=8cm]{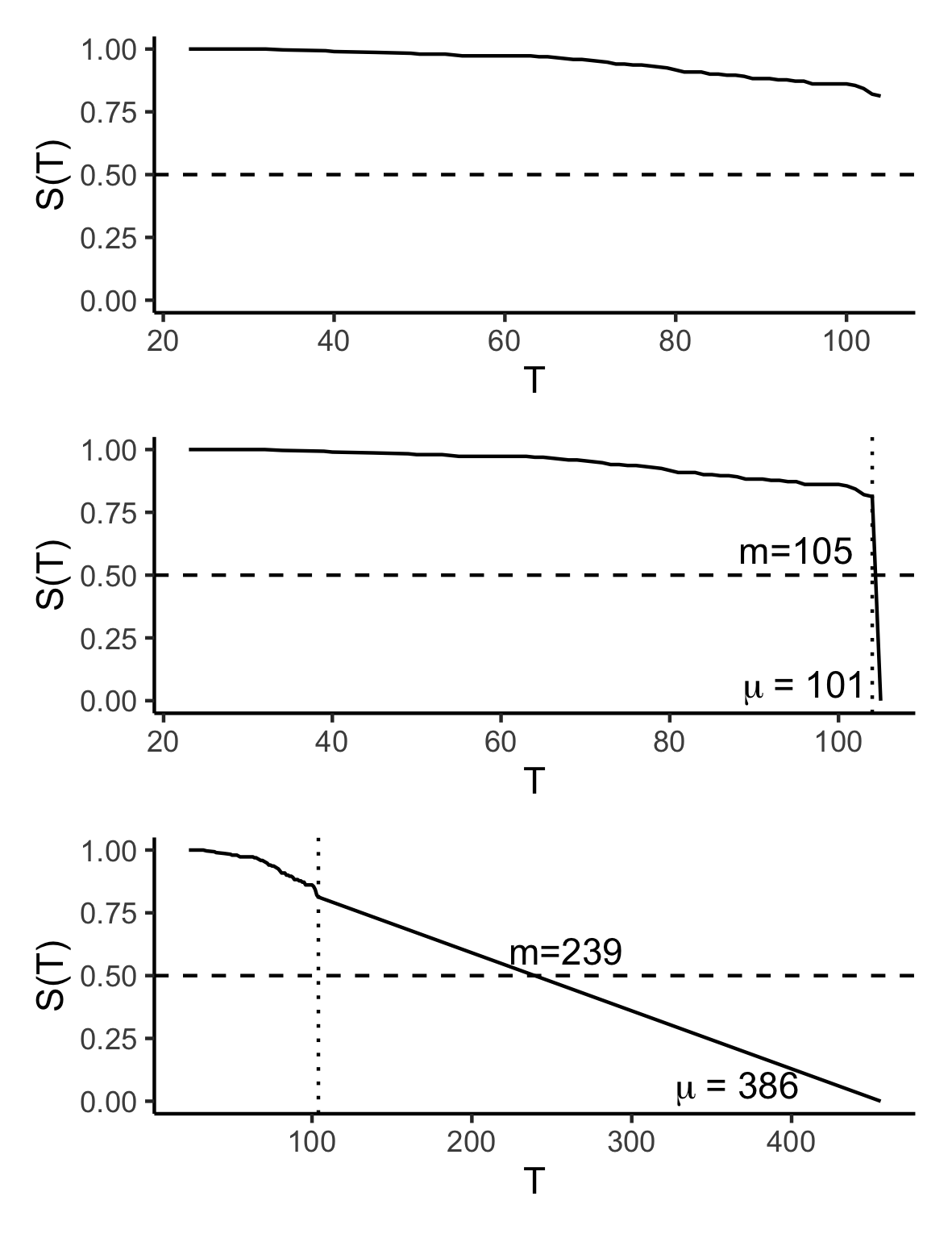}
    \caption{Extrapolation methods to `fix' improper distribution predictions. Top: Kaplan-Meier estimator fit on the
\code{rats}\citep{pkgsurvival} dataset, which results in an improper distribution as $\lim_{T \rightarrow \infty} = 0.81 \neq 0$. Middle: Dropping the survival probability to zero at $T = 105$, just after the study end. Bottom: Dropping the survival probability to zero by linearly extrapolating from first, $(S(T)=1, T = 0)$, and last, $(S(T) = 0.81, T = 104)$, observed survival times. Dashed horizontal lines are drawn at $S(T) = 0.5$ and dotted vertical lines at $T = 104$, where the observed data ends and the extrapolation begins. Median ($m$) and mean ($\mu$) are provided for both extrapolation methods. Both methods result in quantities skewed heavily toward the final extrapolated time. For the `dropping' method the median is exactly at the final time. Linear extrapolation results in probabilities that are unrealistically large.}
    \label{fig:curves}
\end{figure}

Several methods have been considered to extrapolate predictions to fix this problem, such as dropping the last predicted probability to zero either at or just after the last observed time-point \citep{mlr3proba}, or by linear extrapolation from the observed range \citep{Haider2020} (\Cref{fig:curves}). However these methods require unjustifiable assumptions and result in misleading quantities. For example, dropping the survival probability to zero immediately after the study end assumes that all patients (no matter their risk) instantaneously die at the same time, which will skew the distribution mean and median towards the final event time \citep{Haider2020}. The extrapolation method has the opposite problem, if the prediction survival curves are shallow then the extrapolated predictions can easily result in impossible (or at least highly unrealistic) values \See{fig:curves}.

However, we note that summarising a `proper' distribution prediction (i.e. one that doesn't violate the limit properties) by its mean or median will provide a natural relative risk. But this is rarely the case for all predicted distributions in a test set and so the problem remains.

\subsubsection{Expected mortality}
\label{sec:mortality}

The final time-dependent discrimination method estimates

\begin{equation}
    P(\phi_i > \phi_j | T_i < T_j \cap \Delta_i = 1)
\end{equation}

where

\begin{equation}
    \phi_i := \sum_{t \in \calT} -\log \hatS_i(t) = \sum_{t \in \calT} \hatH_i(t)
\end{equation}

and $\hatH_i, \hatS_i$ are the predicted cumulative hazard and survival functions respectively. Summing over the predicted cumulative hazard provides a measure of expected mortality for similar individuals \citep{dataapplied, Ishwaran2008} and a closely related quantity can even be used as measure of calibration \citep{VanHouwelingen2000}.

The advantage of this method is that it requires no model assumptions, nor assumptions about the survival distribution before or after the observed time period, and finally, the reduction method provides an interpretable quantity that is meaningful as a relative risk: the higher the expected mortality, the greater the risk of the event.

\section{Motivating Example}
\label{sec:ex}

We now present a motivating example to make clear why these different concordance measures cannot be directly compared in model evaluation and why it is important to be precise about which method is utilised in model comparison studies. 

\paragraph{Experiment Design}
We split the \code{rats} dataset from \Rstats package \pkg{survival}\citep{pkgsurvival} into a random holdout split with 2/3 of the dataset for training and 1/3 for testing; a seed was set for reproducibility. With the training data we fit a Cox PH (CPH) with package \pkg{survival}, random survival forest (RSF) with package \pkg{ranger}, and gradient boosting machine with C-index optimisation (GBM) \citep{Mayr2014} with package \pkg{mboost} \citep{pkgmboost}. Note that \pkg{ranger} only returns distribution predictions for RSFs and \pkg{mboost} only returns risk predictions.

\paragraph{Evaluation Measures}
We used each model to make predictions on the holdout data. For the Cox PH we made linear predictor predictions with \code{survival::coxph} and additionally distribution predictions with \code{survival::survfit}. We evaluate the discrimination of all possible predictions with: Harrell's C, $C_H$, \citep{Harrell1982} (`Harrell') on the native risk prediction (i.e. returned by package without further user transformation), Uno's C \citep{Uno2011} (`Uno') on the native risk prediction, Antolini's C \citep{Antolini2005} (`Antolini'), $C_H$ computed on the survival probabilities at every predicted time-point, $C_H$ computed on the distribution mean without any extrapolation (`Summary (naive)'), $C_H$ computed on the distribution mean with extrapolation method of dropping to zero just after the final time point (`Summary (extr)'), and $C_H$ computed on the expected mortality (`ExpMort'). For reporting the concordance computed on survival probabilities at each time-point, we reported the time-point which resulted in the maximum $C_H$ for the RSF, the time-point that resulted in the minimum $C_H$ for the RSF, and one randomly sampled time-point. Note that `Harrell' and `Uno' in \Cref{tab:exp} cannot be computed for RSF, which does not return a native risk prediction. Similarly, for the GBM only `Harrell' and `Uno' can be computed as all other methods require a distribution prediction, which does not exist for C-index boosted GBMs.

\paragraph{Results}
The results (\Cref{tab:exp}) indicate how ranking the performance of different algorithms changes depending on the C-index used. The following are examples for how the results in the table could be reported (from most transparent to least):
\begin{enumerate}
    \item CPH is the best performing for distribution predictions under Antolini's C with a C-index of 0.852 compared to RSF's 0.757
    \item RSF is the best performing for distribution predictions under the expected mortality transformation with Harrell's C with a C-index of 0.878 compared to CPH's 0.859.
    \item CPH is the best performing for risk predictions under Uno's C with a C-index of 0.861 compared to GBM's 0.853
    \item RSF is the best performing model with a C-index of 0.897, then CPH with C-index of 0.861 and then GBM with C-index of 0.853.
\end{enumerate}
The first three of these are the clearest as they demonstrate what is being evaluated and how. However, the difference between the first two demonstrates how the result can be chosen by the researcher by selecting one measure over another. The final is clearly the least transparent as it mixes many types of predicted types and evaluation measures to draw conclusions.

\paragraph{Discussion}
These examples demonstrate how simply reporting ``the C-index'' without being more precise can lead to manipulation of results (deliberate or otherwise). For example, the absurdly low values for `Summary (naive)' are a result of attempting to calculate the distribution mean from improper distribution predictions, which is easily possible with \pkg{lifelines} \citep{Davidson-Pilon2019} and \pkg{mlr3proba} \citep{mlr3proba} (the latter has since been updated in light of this problem). Similarly, despite providing a warning in documentation and on usage, \pkg{pec} \citep{pkgpec} still allows concordance evaluation at arbitrary survival points, which could lead to authors reporting the maximum C-index over all time-points (`Prob (max)' in \Cref{tab:exp}).

It is clear that a shift in reporting is required. When a range of C-indices are tabulated as in \Cref{tab:exp} then dishonest reporting (like the final example above) is clear however in practice a range of values is not reported and instead just a vague `C-index'. This problem is analogous to any statistical manipulation, for example p-hacking. The methods of dealing with the problem, `C-hacking', are therefore also the same. Researchers should either: 1) decide at the beginning of an experiment what method they will use for evaluating concordance and state this clearly; or 2) report a wide range of methods depending on utilised software.

\begin{table}[h]
\centering 
\begin{tabular}{cccc} 
\toprule 
\textbf{Method} & \textbf{CPH} & \textbf{RSF} & \textbf{GBM} \\
\midrule
Harrell & $0.859$ & $-$ & 0.831 \\ 
Uno & \textbf{0.861} & $-$ & \textbf{0.853} \\ 
Antolini & $0.852$ & $0.757$ & $-$ \\ 
Prob (min) & $0.500$ & $0.500$ & $-$ \\ 
Prob (max) & $0.859$ & \textbf{0.897} & $-$ \\ 
Prob (rand) & $0.859$ & $0.851$ & $-$ \\ 
Summary (naive) & $0.141$ & $0.104$ & $-$ \\ 
Summary (extr) & $0.859$ & $0.871$ & $-$ \\ 
ExpMort & $0.859$ & $0.878$ & $-$ \\ 
\bottomrule
\end{tabular}
  \caption{Various C-index calculations from different methods and models. Included models are Cox PH (CPH), random survival forest (RSF) and gradient boosting machine with C-index optimisation (CPH). Methods of evaluation are: Harrell's C ($C_H$) on native risk predictions (i.e. predictions from software without further transformation) (`Harrell'), Uno's C on native risk predictions (`Uno'), Antolini's C on distribution predictions (`Antolini'), $C_H$ computed on the predicted survival probability at the time-point that results in the smallest value for RSF (`Prob (min)'), $C_H$ computed on the predicted survival probability at the time-point that results in the largest value for RSF (`Prob (min)'), $C_H$ computed on the predicted survival probability at an arbitrary time-point (`Prob (rand)'), $C_H$ computed on the distribution expectation without any extrapolation (`Summary (naive)'), $C_H$ computed on the distribution expectation after extrapolating by dropping survival probabilities to zero (Fig 2 middle) (`Summary (extr)'), $C_H$ computed on the expected mortality (`ExpMort'). Empty values indicate that the given method is incompatible with the resulting predictions. Values in bold are the maximum C-index for that model.} 
  \label{tab:exp} 
\end{table}

\section{Conclusions}

In this paper, we reviewed the literature for different methods of evaluating survival distribution predictions with methods of concordance. For time-dependent measures, only Antolini's C can be directly applied to distribution predictions. This measure can be utilised to compare the discrimination of multiple models that make distribution predictions however as it cannot be applied to models that make risk predictions, its use in benchmark experiments is more limited. In contrast, methods that reduce a distribution prediction to a risk prediction allow for time-independent discrimination measures to be utilised for any combination of survival models. Of the reviewed `distribution reduction' methods that we found in the literature, the expected mortality method of summing over the cumulative hazard was the most robust as it requires no assumptions about the model or prediction and is therefore applicable to all distribution predictions. Once the distribution is reduced to a risk, any time-independent discrimination measure can be applied (e.g. Harrell's C).

Our motivating example demonstrates why understanding the differences between these methods is so important and how an imprecise statement of methods can lead to simple manipulation of results. We believe this is an important change that must be made from the top-down to ensure it is quickly adopted. Journals should require clear reporting on how c-statistics are computed in survival analysis to ensure fair reporting of results and to avoid `C-hacking'. Furthermore, all open-source software should provide methods to transform distribution to risk predictions, such as the compositions in \cite{mlr3proba}.

In general, researchers should make use of a wide range of evaluation metrics including measures of calibration as well as scoring rules. Whichever metrics are chosen, researchers should be precise about exactly which estimators are utilised and any post-processing of results that was required.


\section*{Competing interests}
There is NO Competing Interest.

\section*{Author contributions statement}

RS conceptualised the article. All authors contributed equally to writing and editing.

\section*{Funding}

AB has been funded by the German Federal Ministry of Education and Research (BMBF) under Grant No. 01IS18036A. The authors of this work take full responsibilities for its content.

\bibliographystyle{abbrvnat}
\bibliography{chacking}

\begin{thebibliography}{43}
\providecommand{\natexlab}[1]{#1}
\providecommand{\url}[1]{\texttt{#1}}
\expandafter\ifx\csname urlstyle\endcsname\relax
  \providecommand{\doi}[1]{doi: #1}\else
  \providecommand{\doi}{doi: \begingroup \urlstyle{rm}\Url}\fi

\bibitem[Aivaliotis et~al.(2021)Aivaliotis, Palczewski, Atkinson, Cade, and
  Morris]{Aivaliotis2021}
G.~Aivaliotis, J.~Palczewski, R.~Atkinson, J.~E. Cade, and M.~A. Morris.
\newblock {A comparison of time to event analysis methods, using weight status
  and breast cancer as a case study}.
\newblock \emph{Scientific Reports}, 11\penalty0 (1):\penalty0 14058, 2021.
\newblock ISSN 2045-2322.
\newblock \doi{10.1038/s41598-021-92944-z}.

\bibitem[Antolini et~al.(2005)Antolini, Boracchi, and Biganzoli]{Antolini2005}
L.~Antolini, P.~Boracchi, and E.~Biganzoli.
\newblock {A time-dependent discrimination index for survival data}.
\newblock \emph{Statistics in Medicine}, 24\penalty0 (24):\penalty0 3927--3944,
  dec 2005.
\newblock ISSN 0277-6715.
\newblock \doi{10.1002/sim.2427}.

\bibitem[Bischl et~al.(2016)Bischl, Lang, Kotthoff, Schiffner, Richter,
  Studerus, Casalicchio, and Jones]{pkgmlr}
B.~Bischl, M.~Lang, L.~Kotthoff, J.~Schiffner, J.~Richter, E.~Studerus,
  G.~Casalicchio, and Z.~M. Jones.
\newblock {mlr: Machine Learning in R}.
\newblock \emph{Journal of Machine Learning Research}, 17\penalty0
  (170):\penalty0 1----5, 2016.
\newblock URL \url{http://jmlr.org/papers/v17/15-066.html
  https://cran.r-project.org/package=mlr}.

\bibitem[Blanche et~al.(2019)Blanche, Kattan, and Gerds]{Blanche2019}
P.~Blanche, M.~W. Kattan, and T.~A. Gerds.
\newblock The c-index is not proper for the evaluation of $t$-year predicted
  risks.
\newblock \emph{Biostatistics}, 20\penalty0 (2):\penalty0 347--357, 2019.
\newblock ISSN 1465-4644.
\newblock \doi{10.1093/biostatistics/kxy006}.

\bibitem[Collins et~al.(2014)Collins, {De Groot}, Dutton, Omar, Shanyinde,
  Tajar, Voysey, Wharton, Yu, Moons, and Altman]{Collins2014}
G.~S. Collins, J.~A. {De Groot}, S.~Dutton, O.~Omar, M.~Shanyinde, A.~Tajar,
  M.~Voysey, R.~Wharton, L.~M. Yu, K.~G. Moons, and D.~G. Altman.
\newblock {External validation of multivariable prediction models: A systematic
  review of methodological conduct and reporting}.
\newblock \emph{BMC Medical Research Methodology}, 14\penalty0 (1):\penalty0
  1--11, 2014.
\newblock ISSN 14712288.
\newblock \doi{10.1186/1471-2288-14-40}.

\bibitem[Cox(1972)]{Cox1972}
D.~R. Cox.
\newblock {Regression Models and Life-Tables}.
\newblock \emph{Journal of the Royal Statistical Society: Series B (Statistical
  Methodology)}, 34\penalty0 (2):\penalty0 187--220, 1972.

\bibitem[Cromb{\'{e}} et~al.(2021)Cromb{\'{e}}, Cousin, Spalato-Ceruso, {Le
  Loarer}, Toulmonde, Michot, Kind, Stoeckle, and Italiano]{Crombe2021}
A.~Cromb{\'{e}}, S.~Cousin, M.~Spalato-Ceruso, F.~{Le Loarer}, M.~Toulmonde,
  A.~Michot, M.~Kind, E.~Stoeckle, and A.~Italiano.
\newblock {Implementing a Machine Learning Strategy to Predict Pathologic
  Response in Patients With Soft Tissue Sarcomas Treated With Neoadjuvant
  Chemotherapy}.
\newblock \emph{JCO Clinical Cancer Informatics}, \penalty0 (5):\penalty0
  958--972, sep 2021.
\newblock \doi{10.1200/CCI.21.00062}.

\bibitem[Davidson-Pilon(2019)]{Davidson-Pilon2019}
C.~Davidson-Pilon.
\newblock lifelines: survival analysis in python.
\newblock \emph{Journal of Open Source Software}, 4\penalty0 (40):\penalty0
  1317, 2019.
\newblock \doi{10.21105/joss.01317}.
\newblock URL \url{https://doi.org/10.21105/joss.01317}.

\bibitem[Fern{\'{a}}ndez et~al.(2016)Fern{\'{a}}ndez, Rivera, and
  Teh]{Fernandez2016}
T.~Fern{\'{a}}ndez, N.~N. Rivera, and Y.~W. Teh.
\newblock {Gaussian Processes for Survival Analysis}.
\newblock \emph{Neural Information Processing Systems}, \penalty0 (Nips), 2016.
\newblock ISSN 10495258.
\newblock URL \url{http://arxiv.org/abs/1611.00817}.

\bibitem[Gensheimer and Narasimhan(2019)]{Gensheimer2019}
M.~F. Gensheimer and B.~Narasimhan.
\newblock {A scalable discrete-time survival model for neural networks}.
\newblock \emph{PeerJ}, 7:\penalty0 e6257, 2019.
\newblock ISSN 2167-8359.

\bibitem[G{\"{o}}nen and Heller(2005)]{Gonen2005}
M.~G{\"{o}}nen and G.~Heller.
\newblock {Concordance Probability and Discriminatory Power in Proportional
  Hazards Regression}.
\newblock \emph{Biometrika}, 92\penalty0 (4):\penalty0 965--970, 2005.

\bibitem[Hadanny et~al.(2022)Hadanny, Shouval, Wu, Gale, Unger, Zahger,
  Gottlieb, Matetzky, Goldenberg, Beigel, and Iakobishvili]{Hadanny2022}
A.~Hadanny, R.~Shouval, J.~Wu, C.~P. Gale, R.~Unger, D.~Zahger, S.~Gottlieb,
  S.~Matetzky, I.~Goldenberg, R.~Beigel, and Z.~Iakobishvili.
\newblock {Machine learning-based prediction of 1-year mortality for acute
  coronary syndrome}.
\newblock \emph{Journal of Cardiology}, jan 2022.
\newblock ISSN 0914-5087.
\newblock \doi{10.1016/j.jjcc.2021.11.006}.

\bibitem[Haider et~al.(2020)Haider, Hoehn, Davis, and Greiner]{Haider2020}
H.~Haider, B.~Hoehn, S.~Davis, and R.~Greiner.
\newblock {Effective ways to build and evaluate individual survival
  distributions}.
\newblock \emph{Journal of Machine Learning Research}, 21\penalty0
  (85):\penalty0 1--63, 2020.
\newblock ISSN 1533-7928.

\bibitem[Harrell et~al.(1982)Harrell, Califf, and Pryor]{Harrell1982}
F.~E. Harrell, R.~M. Califf, and D.~B. Pryor.
\newblock {Evaluating the yield of medical tests}.
\newblock \emph{JAMA}, 247\penalty0 (18):\penalty0 2543--2546, may 1982.
\newblock ISSN 0098-7484.
\newblock \doi{10.1001/jama.1982.03320430047030}.

\bibitem[Heagerty et~al.(2000)Heagerty, Lumley, and Pepe]{Heagerty2000}
P.~J. Heagerty, T.~Lumley, and M.~S. Pepe.
\newblock {Time-Dependent ROC Curves for Censored Survival Data and a
  Diagnostic Marker}.
\newblock \emph{Biometrics}, 56\penalty0 (2):\penalty0 337--344, 2000.
\newblock ISSN 0006-341X.
\newblock \doi{10.1111/j.0006-341X.2000.00337.x}.

\bibitem[Herrmann et~al.(2020)Herrmann, Probst, Hornung, Jurinovic, and
  Boulesteix]{Herrmann2020}
M.~Herrmann, P.~Probst, R.~Hornung, V.~Jurinovic, and A.-L. Boulesteix.
\newblock {Large-scale benchmark study of survival prediction methods using
  multi-omics data}.
\newblock \emph{arXiv preprint arXiv:2003.03621}, 2020.

\bibitem[{Hosmer Jr} et~al.(2011){Hosmer Jr}, Lemeshow, and May]{dataapplied}
D.~W. {Hosmer Jr}, S.~Lemeshow, and S.~May.
\newblock \emph{{Applied survival analysis: regression modeling of
  time-to-event data}}, volume 618.
\newblock John Wiley \& Sons, 2011.
\newblock ISBN 1118211588.

\bibitem[Hothorn et~al.(2020)Hothorn, Buehlmann, Kneib, Schmid, and
  Hofner]{pkgmboost}
T.~Hothorn, P.~Buehlmann, T.~Kneib, M.~Schmid, and B.~Hofner.
\newblock {mboost: Model-Based Boosting}, 2020.
\newblock URL \url{https://cran.r-project.org/package=mboost}.

\bibitem[Ishwaran et~al.(2008)Ishwaran, Kogalur, Blackstone, and
  Lauer]{Ishwaran2008}
B.~H. Ishwaran, U.~B. Kogalur, E.~H. Blackstone, and M.~S. Lauer.
\newblock {Random survival forests}.
\newblock \emph{The Annals of Statistics}, 2\penalty0 (3):\penalty0 841--860,
  2008.
\newblock \doi{10.1214/08-AOAS169}.

\bibitem[Ishwaran and Kogalur(2018)]{pkgrfsrc}
H.~Ishwaran and U.~B. Kogalur.
\newblock {randomForestSRC}, 2018.
\newblock URL \url{https://cran.r-project.org/package=randomForestSRC}.

\bibitem[Johri et~al.(2021)Johri, Mantella, Jamthikar, Saba, Laird, and
  Suri]{Johri2021}
A.~M. Johri, L.~E. Mantella, A.~D. Jamthikar, L.~Saba, J.~R. Laird, and J.~S.
  Suri.
\newblock {Role of artificial intelligence in cardiovascular risk prediction
  and outcomes: comparison of machine-learning and conventional statistical
  approaches for the analysis of carotid ultrasound features and intra-plaque
  neovascularization}.
\newblock \emph{The International Journal of Cardiovascular Imaging},
  37\penalty0 (11):\penalty0 3145--3156, 2021.
\newblock ISSN 1573-0743.
\newblock \doi{10.1007/s10554-021-02294-0}.

\bibitem[Kantidakis et~al.(2020)Kantidakis, Putter, Lancia, de~Boer, Braat, and
  Fiocco]{Kantidakis2020}
G.~Kantidakis, H.~Putter, C.~Lancia, J.~de~Boer, A.~E. Braat, and M.~Fiocco.
\newblock {Survival prediction models since liver transplantation - comparisons
  between Cox models and machine learning techniques}.
\newblock \emph{BMC Medical Research Methodology}, 20\penalty0 (1):\penalty0
  277, 2020.
\newblock ISSN 1471-2288.
\newblock \doi{10.1186/s12874-020-01153-1}.

\bibitem[Kvamme and Borgan(2019)]{Kvamme2019}
H.~Kvamme and {\O}.~Borgan.
\newblock {Continuous and discrete-time survival prediction with neural
  networks}.
\newblock \emph{arXiv preprint arXiv:1910.06724}, 2019.

\bibitem[Kvamme et~al.(2019)Kvamme, Borgan, and Scheel]{Kvamme2019a}
H.~Kvamme, {\O}.~Borgan, and I.~Scheel.
\newblock {Time-to-event prediction with neural networks and Cox regression}.
\newblock \emph{Journal of Machine Learning Research}, 20\penalty0
  (129):\penalty0 1--30, 2019.
\newblock ISSN 1533-7928.

\bibitem[Lee et~al.(2018)Lee, Zame, Yoon, and van~der Schaar]{Lee2018}
C.~Lee, W.~R. Zame, J.~Yoon, and M.~van~der Schaar.
\newblock {Deephit: A deep learning approach to survival analysis with
  competing risks}.
\newblock In \emph{Thirty-Second AAAI Conference on Artificial Intelligence},
  2018.

\bibitem[Loureiro et~al.(2021)Loureiro, Becker, Bauer-Mehren, Ahmidi, and
  Weberpals]{Loureiro2021}
H.~Loureiro, T.~Becker, A.~Bauer-Mehren, N.~Ahmidi, and J.~Weberpals.
\newblock {Artificial Intelligence for Prognostic Scores in Oncology: a
  Benchmarking Study}, 2021.
\newblock URL
  \url{https://www.frontiersin.org/article/10.3389/frai.2021.625573}.

\bibitem[Mayr and Schmid(2014)]{Mayr2014}
A.~Mayr and M.~Schmid.
\newblock {Boosting the concordance index for survival data--a unified
  framework to derive and evaluate biomarker combinations}.
\newblock \emph{PloS one}, 9\penalty0 (1):\penalty0 e84483--e84483, jan 2014.
\newblock ISSN 1932-6203.
\newblock \doi{10.1371/journal.pone.0084483}.
\newblock URL \url{https://pubmed.ncbi.nlm.nih.gov/24400093
  https://www.ncbi.nlm.nih.gov/pmc/articles/PMC3882229/}.

\bibitem[Mogensen et~al.(2014)Mogensen, Ishwaran, and Gerds]{pkgpec}
U.~B. Mogensen, H.~Ishwaran, and T.~A. Gerds.
\newblock {Evaluating Random Forests for Survival Analysis using Prediction
  Error Curves}, 2014.

\bibitem[{Mosquera Orgueira} et~al.(2020){Mosquera Orgueira}, {D{\'{i}}az
  Arias}, {Cid L{\'{o}}pez}, {Peleteiro Ra{\'{i}}ndo}, {Antelo
  Rodr{\'{i}}guez}, {Aliste Santos}, {Alonso Vence}, {Benda{\~{n}}a
  L{\'{o}}pez}, {Abu{\'{i}}n Blanco}, {Bao P{\'{e}}rez}, {Gonz{\'{a}}lez
  P{\'{e}}rez}, {P{\'{e}}rez Encinas}, {Fraga Rodr{\'{i}}guez}, and {Bello
  L{\'{o}}pez}]{MosqueraOrgueira2020}
A.~{Mosquera Orgueira}, J.~{\'{A}}. {D{\'{i}}az Arias}, M.~{Cid L{\'{o}}pez},
  A.~{Peleteiro Ra{\'{i}}ndo}, B.~{Antelo Rodr{\'{i}}guez}, C.~{Aliste Santos},
  N.~{Alonso Vence}, {\'{A}}.~{Benda{\~{n}}a L{\'{o}}pez}, A.~{Abu{\'{i}}n
  Blanco}, L.~{Bao P{\'{e}}rez}, M.~S. {Gonz{\'{a}}lez P{\'{e}}rez}, M.~M.
  {P{\'{e}}rez Encinas}, M.~F. {Fraga Rodr{\'{i}}guez}, and J.~L. {Bello
  L{\'{o}}pez}.
\newblock {Improved personalized survival prediction of patients with diffuse
  large B-cell Lymphoma using gene expression profiling}.
\newblock \emph{BMC Cancer}, 20\penalty0 (1):\penalty0 1017, 2020.
\newblock ISSN 1471-2407.
\newblock \doi{10.1186/s12885-020-07492-y}.

\bibitem[P\"olsterl(2020)]{pkgsksurvival}
S.~P\"olsterl.
\newblock {scikit-survival: A Library for Time-to-Event Analysis Built on Top
  of scikit-learn}.
\newblock \emph{Journal of Machine Learning Research}, 21\penalty0
  (212):\penalty0 1----6, 2020.
\newblock URL \url{http://jmlr.org/papers/v21/20-729.html}.

\bibitem[Potapov et~al.(2012)Potapov, Adler, and Schmid]{pkgsurvauc}
S.~Potapov, W.~Adler, and M.~Schmid.
\newblock {survAUC: Estimators of prediction accuracy for time-to-event data.},
  2012.

\bibitem[Rahman et~al.(2017)Rahman, Ambler, Choodari-Oskooei, and
  Omar]{Rahman2017}
M.~S. Rahman, G.~Ambler, B.~Choodari-Oskooei, and R.~Z. Omar.
\newblock {Review and evaluation of performance measures for survival
  prediction models in external validation settings}.
\newblock \emph{BMC Medical Research Methodology}, 17\penalty0 (1):\penalty0
  1--15, 2017.
\newblock ISSN 14712288.
\newblock \doi{10.1186/s12874-017-0336-2}.

\bibitem[Schwarzer et~al.(2000)Schwarzer, Vach, and Schumacher]{Schwarzer2000}
G.~Schwarzer, W.~Vach, and M.~Schumacher.
\newblock {On the misuses of artificial neural networks for prognostic and
  diagnostic classification in oncology}.
\newblock \emph{Statistics in Medicine}, 19\penalty0 (4):\penalty0 541--561,
  feb 2000.
\newblock ISSN 0277-6715.
\newblock
  \doi{10.1002/(SICI)1097-0258(20000229)19:4<541::AID-SIM355>3.0.CO;2-V}.

\bibitem[Sonabend et~al.(2021)Sonabend, Kir{\'{a}}ly, Bender, Bischl, and
  Lang]{mlr3proba}
R.~Sonabend, F.~J. Kir{\'{a}}ly, A.~Bender, B.~Bischl, and M.~Lang.
\newblock {mlr3proba: An R Package for Machine Learning in Survival Analysis}.
\newblock \emph{Bioinformatics}, feb 2021.
\newblock ISSN 1367-4803.
\newblock \doi{10.1093/bioinformatics/btab039}.
\newblock URL \url{https://cran.r-project.org/package=mlr3proba}.

\bibitem[Spooner et~al.(2020)Spooner, Chen, Sowmya, Sachdev, Kochan, Trollor,
  and Brodaty]{Spooner2020}
A.~Spooner, E.~Chen, A.~Sowmya, P.~Sachdev, N.~A. Kochan, J.~Trollor, and
  H.~Brodaty.
\newblock {A comparison of machine learning methods for survival analysis of
  high-dimensional clinical data for dementia prediction}.
\newblock \emph{Scientific Reports}, 10\penalty0 (1):\penalty0 20410, 2020.
\newblock ISSN 2045-2322.
\newblock \doi{10.1038/s41598-020-77220-w}.

\bibitem[Therneau(2015)]{pkgsurvival}
T.~M. Therneau.
\newblock {A Package for Survival Analysis in S}, 2015.
\newblock URL \url{https://cran.r-project.org/package=survival}.

\bibitem[Uno et~al.(2011)Uno, Cai, Pencina, D'Agostino, and Wei]{Uno2011}
H.~Uno, T.~Cai, M.~J. Pencina, R.~B. D'Agostino, and L.~J. Wei.
\newblock {On the C-statistics for Evaluating Overall Adequacy of Risk
  Prediction Procedures with Censored Survival Data}.
\newblock \emph{Statistics in Medicine}, 30\penalty0 (10):\penalty0 1105--1117,
  2011.
\newblock ISSN 02776715.
\newblock \doi{10.1002/sim.4154}.

\bibitem[{Van Belle} et~al.(2007){Van Belle}, Pelckmans, Suykens, and {Van
  Huffel}]{VanBelle2007}
V.~{Van Belle}, K.~Pelckmans, J.~A. Suykens, and S.~{Van Huffel}.
\newblock {Support Vector Machines for Survival Analysis}.
\newblock In \emph{In Proceedings of the Third International Conference on
  Computational Intelligence in Medicine and Healthcare}, number~1, 2007.
\newblock \doi{10.1016/j.microrel.2005.05.002}.

\bibitem[{Van Houwelingen}(2000)]{VanHouwelingen2000}
H.~C. {Van Houwelingen}.
\newblock {Validation, calibration, revision and combination of prognostic
  survival models}.
\newblock \emph{Statistics in Medicine}, 19\penalty0 (24):\penalty0 3401--3415,
  2000.
\newblock ISSN 02776715.
\newblock \doi{10.1002/1097-0258(20001230)19:24<3401::AID-SIM554>3.0.CO;2-2}.

\bibitem[Wright and Ziegler(2017)]{pkgranger}
M.~N. Wright and A.~Ziegler.
\newblock {ranger: A Fast Implementation of Random Forests for High Dimensional
  Data in C++ and R}.
\newblock \emph{Journal of Statistical Software}, 77\penalty0 (1):\penalty0
  1----17, 2017.

\bibitem[Zhang et~al.(2021)Zhang, Wong, Mann, Muller, and Yang]{Zhang2021}
Y.~Zhang, G.~Wong, G.~Mann, S.~Muller, and J.~Y.~H. Yang.
\newblock {SurvBenchmark: comprehensive benchmarking study of survival analysis
  methods using both omics data and clinical data}.
\newblock \emph{bioRxiv}, page 2021.07.11.451967, jan 2021.
\newblock \doi{10.1101/2021.07.11.451967}.

\bibitem[Zhao and Feng(2020)]{Zhao2020}
L.~Zhao and D.~Feng.
\newblock {DNNSurv: Deep Neural Networks for Survival Analysis Using Pseudo
  Values}.
\newblock 2020.
\newblock URL \url{https://arxiv.org/abs/1908.02337}.

\bibitem[Zhong and Tibshirani(2019)]{Zhong2019}
C.~Zhong and R.~Tibshirani.
\newblock {Survival analysis as a classification problem}.
\newblock sep 2019.
\newblock URL \url{http://arxiv.org/abs/1909.11171}.

\end{thebibliography}

\end{document}